\documentclass{article}

\usepackage[preprint]{neurips_2025}

\usepackage[utf8]{inputenc}
\usepackage[T1]{fontenc}
\usepackage{hyperref}
\usepackage{url}
\usepackage{booktabs}
\usepackage{amsfonts}
\usepackage{amsmath}
\usepackage{nicefrac}
\usepackage{microtype}
\usepackage{xcolor}
\usepackage{enumitem}
\usepackage{graphicx}
\usepackage{multirow}

\graphicspath{{figures/}{../results/figures/}}

\title{Deployment Decision Reliability:\\
A Generalizability-Theory Framework for Sizing Long-Horizon Agent Evaluations\thanks{Code, data loaders, results, and reproducibility instructions: \url{https://github.com/vasundras/agent-reliability-engineering-lab}}}

\author{%
  Vasundra Srinivasan \\
  Independent Researcher, Stanford School of Engineering
}

\begin{document}

\maketitle

\begin{abstract}
Enterprise practitioners read agent leaderboards as if they ranked
\emph{agent capability}. We show, across three open agent-trace
benchmarks (TheAgentCompany, $\tau^2$-bench, and AppWorld), that the
agent main effect accounts for less than 3\% of total variance in
every dataset and check type, while the agent-by-task interaction
accounts for 7--23\%. Leaderboards rank \emph{specialization}, not
capability. We arrive at this through a four-facet Generalizability
Theory variance decomposition, fit with three estimators (Henderson
Method-I, REML via lme4, and a Bayesian binomial GLMM on an NVIDIA
L4 GPU) that agree to three decimal places. Four further findings
sharpen what the leaderboard is hiding. First, aggregate reliability
collapses on the hardest task quartile: $E\rho^2$ on $\tau^2$
\texttt{action\_checks} falls from $0.752$ to $0.000$. Second,
training-cell reliability \emph{negatively} correlates with held-out
reliability ($r = -0.90$ on $\tau^2$), meaning the designs that look
most reliable replicate worst. Third, population-level
diagnostics transfer across enterprise benchmarks (capability-gap ratio
stable at $0.35$--$0.40$) but per-family agent rankings invert. Fourth,
on the MAST failure taxonomy, trace-level mode profiles are idiosyncratic
(MAE $= 0.261$) while cell-level profiles generalise (MAE $= 0.056$,
$r = 0.83$). We package these into \textbf{Deployment Decision
Reliability (DDR)}, a one-page reporting discipline that turns the
variance-component table into five decisions an enterprise buyer can
defend. All code, data loaders, fit artifacts, and reproducibility
instructions are released under an open-source license at
\url{https://github.com/vasundras/agent-reliability-engineering-lab}.
\end{abstract}

\section{Introduction}

When an enterprise practitioner selects one agent over another for a
production workflow, the decision rests on a benchmark score that compresses
many distinct sources of measurement variance into a single number. The
practitioner cannot tell whether the observed gap between two agents
originates in the agents themselves, in the task mix the benchmark happens
to sample, in the step at which the harness cuts each trajectory, in the
failure category the evaluator surfaces, or in interactions among these
sources. More practically, the same score cannot say how many tasks the
practitioner needs in a production evaluation to reproduce the benchmark
gap at a stated confidence, how that requirement changes if the production
task mix is harder than the benchmark average, or how it changes when
inference cost is a binding constraint.

This is a measurement problem, and psychometrics has treated it for decades.
Generalizability Theory~\citep{cronbach1972} (G-theory) decomposes total
observed variance into a signal component attributable to the object of
measurement (here, the agent) and interpretable noise components
corresponding to each design facet (task, step, error category) and to
their pairwise and higher-order interactions. The companion Decision Study
(D-study) then translates those variance components into a budget frontier:
the minimum number of tasks, steps, and judges needed to reach a target
generalizability coefficient $E\rho^2 \geq 0.85$ for relative ranking
decisions, or a dependability $\Phi \geq 0.90$ for absolute-threshold
decisions such as SLA gates.

G-theory already has an evaluation-sizing lineage outside classical
psychometrics: it has been used to ask how many topics an
information-retrieval test collection needs for a reliable system
ranking~\citep{urbano2013reliability}, to compare human and LLM
raters in large-scale writing assessment~\citep{song2025autoscoring},
and to correct hidden measurement error in single-shot LLM annotation
pipelines~\citep{messing2026tee}. What that lineage has not yet
reached is the setting where the object of measurement is a
multi-step agent: trajectories introduce step position and error
category as facets in their own right, and deployment decisions
introduce cost and difficulty constraints that test-collection sizing
never faced. We extend the machinery to this setting and propose
treating the question \emph{what evaluation design supports this
deployment decision?} as a first-class measurement-design question. We
frame the resulting combined output as \textbf{Deployment Decision
Reliability (DDR)}: the property that a given evaluation supports a
deployment-comparison decision at a stated reliability target. DDR is not
a new metric. It is a discipline that combines the variance-component
table, the D-study budget frontier, a cost-adjusted reliability index, a
difficulty-conditional reliability profile, and an explicit
capability-ceiling decomposition into one unified report for practitioners.

\paragraph{Contributions.}
This paper makes one empirical claim and one framework contribution.

\emph{Empirical.} Across three open enterprise agent benchmarks
(TheAgentCompany, $\tau^2$-bench, AppWorld) and one auxiliary failure-mode
dataset (MAST/MAD~\citep{cemri2025mast}), we report five findings that
revise the default practitioner reading of agent leaderboards:

\begin{enumerate}[leftmargin=*,nosep]
  \item \textbf{Capability ceiling.} The agent main effect $\sigma^2_a$
        is below 3\% of total variance in every dataset and check type;
        the agent-by-task interaction is 4--13$\times$ larger. Leaderboard
        rank order reflects \emph{specialization} across tasks, not a
        uniform capability advantage.
  \item \textbf{Hard-task collapse.} Aggregate $E\rho^2$ does not predict
        hard-quartile $E\rho^2$ (e.g., $0.752 \to 0.000$ on $\tau^2$
        \texttt{action\_checks}). When the production task mix is harder
        than the benchmark mix, the headline reliability number provides
        no useful guarantee.
  \item \textbf{Held-out reversal.} Training-cell $E\rho^2$ correlates
        \emph{negatively} with held-out $E\rho^2$ across 50 random splits
        ($r = -0.90$ on $\tau^2$). The training-cell projection
        systematically overstates the reliability a replication will
        recover.
  \item \textbf{Aggregate-versus-family asymmetry.} The capability-gap
        ratio is stable at $0.35$--$0.40$ across TheAgentCompany and
        AppWorld, but per-family rankings invert
        ($\hat{\rho} = -0.50$, $n = 3$). Population diagnostics transfer;
        family-level claims do not.
  \item \textbf{Variance $\neq$ frequency.} Per-category $\sigma^2_{a:t}$
        and per-category failure-activation rates rank orthogonally
        ($\hat{\rho} = -0.20$ on MAST, $n = 4$, descriptive). The
        categories where agents differ are not the categories where
        failures appear.
\end{enumerate}

\emph{Framework.} We introduce \textbf{Deployment Decision Reliability
(DDR)}, a one-page reporting discipline that turns the
variance-component table into five decisions an enterprise buyer can
defend: how many tasks to sample, which difficulty bucket to target,
which cost band to constrain on, which holdout protocol to demand from
a benchmark vendor, and how to read a failure-mode taxonomy. DDR is not
a new metric. It is a checklist that exposes the five findings above
directly to the procurement decision.

\section{Related Work}

\paragraph{Statistical rigor for LLM evaluation.}
A recent line of work quantifies the uncertainty that leaderboards
omit. \citet{miller2024errorbars} derives clustered standard errors
and power analyses for benchmark comparisons;
\citet{madaan2024variance} measure variance across seeds and training
checkpoints on standard benchmarks; and \citet{heineman2025signal}
decompose benchmark signal-to-noise ratios to select evaluations for
model development. These approaches quantify \emph{how much} noise an
evaluation carries. G-theory complements them by decomposing
\emph{where} the noise lives---agent, task, step, error category, or
their interactions---and by supplying, through the D-study, the
evaluation-sizing frontier that a confidence interval alone cannot.

\paragraph{G-theory as an evaluation-sizing instrument.}
Applying G-theory to system evaluation is not itself new.
\citet{urbano2013reliability} used variance decomposition and D-study
projection to size information-retrieval test collections---``how
many topics does a reliable ranking need?''---the single-facet
ancestor of our question, and \citet{urbano2016bias} studied the bias
of those G-theory estimates under violated statistical assumptions
via stochastic simulation. In LLM settings, G-theory has so far
assessed models as \emph{raters}: \citet{song2025autoscoring} compare
human and LLM scoring consistency in large-scale writing assessment.
The closest direct methodological precedent is the Total Evaluation
Error (TEE) framework~\citep{messing2026tee}, which integrates total
survey error and G-theory for single-shot LLM annotation pipelines.
None of these treats a multi-step agent as the object of measurement.
We extend the lineage to a four-facet crossed design in which step
position and error category are first-class variance sources, and add
the cost-adjusted, difficulty-conditional, and held-out diagnostics
that deployment decisions require and test-collection sizing did not.

\paragraph{Composite scoring and psychometric evaluation.}
\citet{liang2022holistic} established the multi-axis composite-score
paradigm with HELM, aggregating seven axes via expert-set weights;
tinyBenchmarks~\citep{polo2024tinybench} applied Item Response Theory
to LLM evaluation. Our approach replaces expert weighting with
empirical variance decomposition, so each axis's weight reflects the
fraction of total measurement variance it explains; G-theory also
generalises IRT's unidimensional latent ability to the multi-faceted
crossed designs that agent evaluation requires.
\citet{ilic2024evidence} find a strong general factor across 591 LLMs
on static cognitive tests via factor analysis; our
$\sigma^2_a \approx 0$ finding does not contradict theirs, but shows
that on \emph{agentic} surfaces the discriminating signal migrates
from a general factor into the agent-by-task interaction.
\citet{ndzomga2026efficient} gives an IRT-guided answer to evaluation
sizing---filter to mid-difficulty tasks---where our D-study derives
the task budget from the full variance decomposition.

\paragraph{Reliability metrics for agent benchmarks.}
$\tau$-bench~\citep{yao2024taubench} introduced \texttt{pass}$^k$, the
probability that all $k$ repeated trials succeed, as a
reliability-oriented metric distinct from one-shot pass rate. In
G-theory terms \texttt{pass}$^k$ is a single-facet (trial) design; our
four-facet crossed design generalises it to model task, step, and
error-category noise simultaneously and to derive the D-study budget
frontier across facets jointly. \citet{zhang2026harness} document that
agent scores shift substantially with the evaluation harness while
rankings only partially persist---a qualitative account of the
harness-as-variance-source effect whose magnitude our
$\sigma^2_{a:t}$ estimates quantify.

\paragraph{Failure-mode taxonomy.}
MAST~\citep{cemri2025mast} provides a 14-mode failure taxonomy and the
MAD dataset derived from 1,642 multi-agent traces across seven frameworks.
Their work classifies failures given a trace; ours measures how those
classifications vary as a function of agent, task, and judgment source,
and uses the MAST taxonomy to define the error-category facet in our
sensitivity analysis.

\paragraph{Step-level failure attribution.}
\citet{zhang2025who} annotated 184 failed multi-agent traces with
\texttt{mistake\_agent} and \texttt{mistake\_step} labels, reporting
that state-of-the-art reasoning models achieve only 14.2\% step-level
attribution accuracy. Who\&{}When frames failure attribution as a
classification problem; our G-study frames it as a measurement problem,
asking not ``which step failed?'' but ``how much of the observed variance
is attributable to step position, and is that variance stable enough to
support a deployment decision?''

\section{Methods}

\paragraph{Measurement design.}
We treat each observation as a binary success indicator $Y_{atse} \in
\{0, 1\}$ at step $s$ for agent $a$ on task $t$ with error-category label
$e$. The four-facet fully crossed G-study decomposes the expected score as:
\begin{equation}
  Y_{atse} = \mu + \nu_a + \nu_t + \nu_s + \nu_e
  + \nu_{at} + \nu_{as} + \nu_{ae} + \nu_{ts} + \nu_{te} + \nu_{se}
  + \varepsilon_{atse},
  \label{eq:gstudy}
\end{equation}
where each $\nu$ term is a zero-mean random effect with variance
$\sigma^2_{\cdot}$ and $\varepsilon_{atse}$ is the residual. The object
of measurement is \emph{agent}, so $\sigma^2_a$ is the signal of interest.
We omit three-way and four-way interactions because their cell counts are
uneven across datasets and pilot fits returned zero estimates for all
higher-order components.

\paragraph{Estimators.}
We use three estimators whose agreement provides a cross-validation of the
variance-component estimates. Henderson Method-I~\citep{henderson1953} is
a closed-form ANOVA-based estimator that may return negative values for
ill-identified components; we floor these at zero in tables and flag them
in text. Restricted maximum likelihood (REML) via lme4~\citep{lme4} is the
canonical frequentist estimator for unbalanced Gaussian random-effects
models. Bayesian binomial GLMM via bambi/numpyro~\citep{bambi,numpyro}
is the principled estimator for binary outcomes; we report posterior
medians and 95\% credible intervals (CrIs) for each $\sigma$ on the logit
scale, converting to $\sigma^2$ for comparability with the other
estimators.

\paragraph{Reliability coefficients.}
Given the estimated variance components, we compute two coefficients. The
generalizability coefficient $E\rho^2 = \sigma^2_a / (\sigma^2_a +
\sigma^2_\delta)$ quantifies relative reliability for ranking decisions,
where $\sigma^2_\delta$ is the relative-error variance, defined as
the sum of all interaction terms involving the agent, divided by
their respective sample sizes in the design. The dependability coefficient $\Phi = \sigma^2_a /
(\sigma^2_a + \sigma^2_\Delta)$ quantifies absolute reliability for
threshold decisions, where $\sigma^2_\Delta$ additionally includes the
main effects of non-agent facets divided by their sample sizes.

\paragraph{Capability-ceiling decomposition.}
Our pre-analysis hypothesis was that $\sigma^2_a$ would dominate. When
the first round of fits returned $\sigma^2_a \approx 0$ on every dataset,
we promoted this to an explicit diagnostic. The \emph{capability-gap
ratio} $\sigma^2_a / (\sigma^2_a + \sigma^2_{a:t})$ measures what share of
agent-related variance lives in the main effect versus the agent-by-task
interaction. When this ratio is near zero, agents do not differ in overall
capability on the evaluation surface; they differ in which tasks they
handle well, a specialisation pattern rather than a dominance
ordering.

\paragraph{Cost-aware reliability (RPD).}
Reliability per Dollar is $\mathrm{RPD}(a, B) = E\rho^2(a, B) / \bar{C}(a, B)$,
where $\bar{C}$ denotes the mean inference cost per task observation.
Costs are derived from $\tau^2$'s native per-simulation token counts joined
against publicly reported model pricing.
RPD ranks are reported alongside accuracy ranks to reveal which agents are
favoured or penalised by cost-aware procurement.

\paragraph{Difficulty-conditional reliability.}
Tasks are stratified into quartiles by their mean success rate across all
agents ($Q_1$ = easiest, $Q_4$ = hardest). A separate G-study is fit
within each stratum. We report $E\rho^2_{Q_k}$ and the gap
$E\rho^2_{\text{agg}} - E\rho^2_{Q_4}$ as the measurement cost of using
aggregate reliability to support claims about hard-task deployments.

\paragraph{Hold-out generalization test.}
We perform 50 random 70/30 splits stratified by (agent, task) cell. For
each split we fit the G-study on training cells, project the implied
$E\rho^2$ for the observed design, and compare it to the empirical
$E\rho^2$ on held-out cells. A positive Pearson correlation between
projected and held-out reliability across 50 splits would indicate that
training-cell fits generalise; a negative correlation is a DDR-cautionary
finding requiring practitioners to report a held-out reliability estimate
rather than reading the training-cell projection as ground truth.

\paragraph{Cross-dataset transfer.}
To test whether DDR diagnostics generalise across enterprise domains, we
rank agent families by mean success on each dataset and compute Spearman's
$\rho$ between datasets within shared families. We also compare the
population capability-gap ratio across datasets, which tests whether
the aggregate diagnostic transfers across domains even when per-family
rankings do not.

\paragraph{Failure-mode analysis.}
We conduct two complementary analyses using the MAST/MAD dataset. First, a
within-MAST G-study applies the design in Equation~\ref{eq:gstudy} to MAD
with the 14 native failure modes as the ``step'' facet, followed by a
70/30 trace-level hold-out (50 random splits) that uses the
cell-by-mode mean activation rate from training to predict held-out trace
indicators, evaluating generalisability of the 14-mode profile. Second, a
cross-dataset descriptive contrast maps the 14 MAST modes onto our
four-category taxonomy and compares $\tau^2$ $\sigma^2_{a:t}$ per category
against MAST per-category activation shares.

\section{Data and Setup}

We use three primary agent-trace datasets and one auxiliary failure-mode
annotation surface. Our pre-analysis plan named AndroidWorld and AppWorld
as the primary G-study datasets. Phase 1 trace-data hunting revised this:
AndroidWorld and OdysseyBench release no public step-level trace data (the
plan's risk register flagged live-emulator collection as a 3--4 day
commitment to be cut if it slipped), and we cut it. The replacement
datasets (TheAgentCompany~\citep{xu2024agentcompany},
$\tau^2$-bench~\citep{tau2}, and AppWorld~\citep{trivedi2024appworld})
meet or exceed the original choices on every criterion the plan
specified:
public availability, ground-truth state inspection rather than LLM judges,
step-level labels, multiple agent harnesses, and enterprise or long-horizon
domain coverage. AppWorld was retained from the original plan; the
MAST/MAD dataset~\citep{cemri2025mast} was added as the failure-mode
auxiliary surface.

\paragraph{TheAgentCompany.}
175 enterprise tasks spanning GitLab, OwnCloud, Plane, and RocketChat,
each evaluated at 1--6 environment-state checkpoints with partial
credit. Ground-truth state inspection removes the LLM-judge confound.
Filtered to 17 agents $\times$ 175 task bases $\times$ 1--3 checkpoints
$\times$ $\{$text, image$\}$, yielding $7{,}783$ rows.

\paragraph{$\tau^2$-bench.}
Customer-service workflows in airline, retail, and telecom domains;
three frontier agents (Claude~3.7 Sonnet, GPT-4.1, o4-mini) running 4
trials per task across 50 tasks per domain. The reward schema
decomposes correctness into five check types (\texttt{db\_check},
\texttt{action\_checks}, \texttt{env\_assertions},
\texttt{communicate\_checks}, \texttt{nl\_assertions}); we fit one
G-study per check type because each measures a distinct dimension.

\paragraph{AppWorld.}
750 enterprise-app tasks, $\sim$8 unit tests per task. We merged the
canonical \texttt{appworld download experiment-outputs} bundle (14
agent configurations) with four additional agents pulled from the
public leaderboard (Cuga, Loop, Alibaba AgentRL Qwen3-14B, demo-augmented
ReAct), totalling 18 agent configurations $\times$ 2 splits and
$79{,}650$ unit-test rows.

\paragraph{MAD/MAST.}
1,242 LLM-judged traces with 14 binary failure-mode labels, drawn from
7 multi-agent systems $\times$ 7 benchmarks (9 populated cells), plus a
19-trace human-annotated subset for validity check.

\paragraph{Compute environment.}
REML used R~4.5 with lme4~2.0.1; Bayesian fits used Python~3.11 with
bambi/pymc/numpyro on a spot NVIDIA L4 (g2-standard-4, 23\,GB VRAM).
The full sweep of 10 fits (4 chains $\times$ 1000 draws) completed in
$\approx$70 minutes on GPU and 4 hours on CPU, with posteriors
agreeing to 3--4 decimal places.

\section{Results}\label{sec:results}

We organise the results around five DDR diagnostics. Each subsection
reports the empirical finding, its measurement interpretation, and the
practitioner implication. Table~\ref{tab:headline-vc} provides the
anchor variance-component estimates; the remaining diagnostics build on
those estimates or extend them to new analysis surfaces.

\subsection{Variance decomposition and capability ceiling}

Table~\ref{tab:headline-vc} presents the REML variance components for
each (dataset, check-type) cell. The headline result holds uniformly: the
agent main effect $\sigma^2_a$ is below 3\% of total variance in every
cell and is exactly zero on $\tau^2$ \texttt{db\_check} and
\texttt{nl\_assertions}. By contrast, the agent-by-task interaction
$\sigma^2_{a:t}$ accounts for 7.5--12.5\% of total variance, one to two
orders of magnitude larger than $\sigma^2_a$. The capability-gap ratio
$\sigma^2_a / (\sigma^2_a + \sigma^2_{a:t})$ is below 0.001 on
TheAgentCompany and below 0.19 on every $\tau^2$ cell. The pre-analysis
hypothesis that $\sigma^2_a$ would dominate is rejected; instead, agents
have different specialisation patterns across tasks rather than different
overall capability levels.

\begin{table}[t]
\centering
\small
\caption{Headline variance decomposition (REML). $\sigma^2_a$: agent main
  effect; $\sigma^2_{a:t}$: agent-by-task interaction. Gap ratio:
  $\sigma^2_a / (\sigma^2_a + \sigma^2_{a:t})$. $E\rho^2$: aggregate
  generalizability coefficient. All figures as percentage of total
  variance for $\sigma^2$ columns.}
\label{tab:headline-vc}
\setlength{\tabcolsep}{6pt}
\begin{tabular}{llcccc}
\toprule
Dataset & Check type & $\sigma^2_a$~(\%) & $\sigma^2_{a:t}$~(\%) & Gap ratio & $E\rho^2$ \\
\midrule
TheAgentCompany & all           & ${<}0.001$ & 12.51 & ${<}0.001$ & 0.986 \\
\midrule
\multirow{4}{*}{$\tau^2$-bench}
  & action\_checks     & 1.74  & 7.54  & 0.187 & 0.752 \\
  & db\_check          & 0.00  & 12.35 & 0.000 & 0.899 \\
  & env\_assertions    & 2.34  & 10.98 & 0.176 & 0.316 \\
  & nl\_assertions     & 0.00  & 11.43 & 0.000 & 0.000 \\
\bottomrule
\end{tabular}
\end{table}

The Bayesian binomial GLMM estimates confirm this pattern while providing
principled uncertainty quantification for binary outcomes. On
TheAgentCompany, the posterior for $\sigma_a$ on the logit scale has
median 2.19 with 95\% CrI $[0.14, 3.87]$, a wide interval that reflects
the sparse identification of the agent main effect with 17 agents. The
posterior for $\sigma_{a:t}$ is $2.26$ (95\% CrI $[1.99, 2.54]$),
sharply identified and consistent with REML. The key measurement
implication is that the interaction is reliably estimated while the main
effect is not, which means leaderboard rankings based on aggregate success
rate carry substantial measurement uncertainty. All REML and CPU posteriors
agree to 3--4 decimal places with the GPU posteriors, providing an
independent estimator check.

\subsection{Cost-aware reliability}

On $\tau^2$ \texttt{action\_checks}, the three frontier agents rank by
accuracy in descending order as Claude~3.7~Sonnet (83\%), GPT-4.1
(81\%), o4-mini (78\%). Under RPD the order becomes
Claude~3.7~Sonnet, o4-mini, GPT-4.1: o4-mini advances from third to
second because its slightly lower accuracy is offset by its
substantially lower per-turn inference cost. The accuracy ranking and
the RPD ranking disagree on every $\tau^2$ check type except
\texttt{db\_check}, so the headline accuracy number alone cannot
recover a procurement decision under a cost constraint
(\texttt{ext1\_rpd\_ranking.csv}).

\subsection{Difficulty-conditional reliability}

\begin{figure}[t]
\centering
\includegraphics[width=\textwidth]{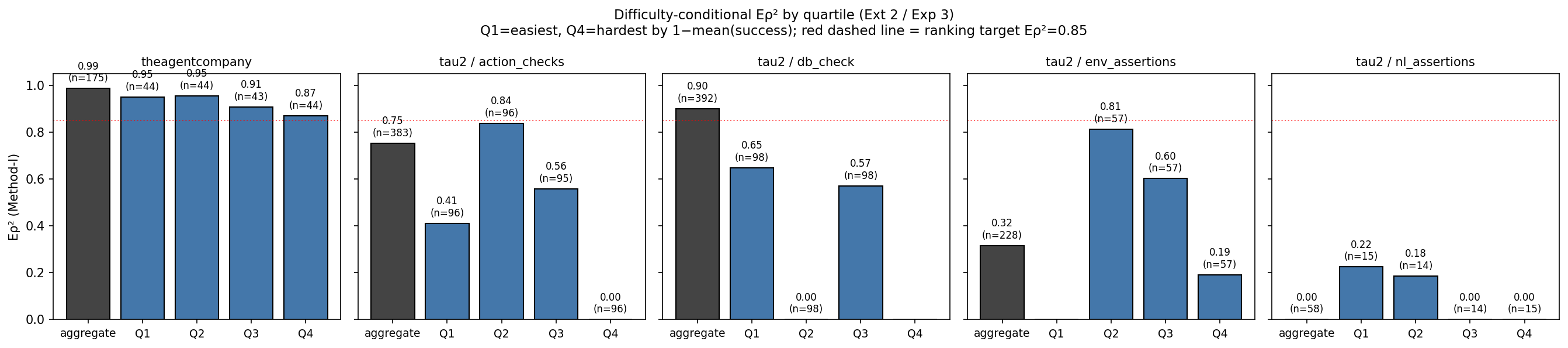}
\caption{Difficulty-conditional $E\rho^2$ across $\tau^2$ check types and
TheAgentCompany. Quartiles are defined on mean task success ($Q_1$ =
easiest, $Q_4$ = hardest). The aggregate coefficient (rightmost bar in
each panel) systematically overstates reliability on the hard quartile;
on $\tau^2$ \texttt{action\_checks} and \texttt{nl\_assertions}, $E\rho^2$
collapses to zero in $Q_4$ because $\sigma^2_a = 0$ among the hardest
tasks.}
\label{fig:difficulty}
\end{figure}

The aggregate $E\rho^2$ on TheAgentCompany is $0.986$, which appears to
indicate near-perfect rank reliability. Conditioning on the hardest task
quartile (Q4, mean success rate $0.066$), $E\rho^2_{Q_4}$ falls to $0.869$,
a reduction of $0.117$. On $\tau^2$ \texttt{action\_checks} the effect is
more severe: aggregate $E\rho^2 = 0.752$ collapses to $0.000$ in Q4,
because $\sigma^2_a = 0$ among the hardest tasks in this check type. A
practitioner whose production deployment skews toward the hardest quartile
of tasks cannot read the aggregate $E\rho^2 = 0.752$ and assume it
generalises to their setting; on $\tau^2$ \texttt{action\_checks}, there
is no aggregate-level agent reliability to exploit in the hard stratum.
This finding motivates reporting difficulty-conditional $E\rho^2$ whenever
the production task-difficulty distribution is known to differ from the
benchmark distribution. Figure~\ref{fig:difficulty} visualises the
collapse across $\tau^2$ check types.

\subsection{Hold-out generalization}

Across 50 random 70/30 training/test splits, variance components estimated
on training cells systematically overestimate the reliability available on
held-out cells. On $\tau^2$ \texttt{action\_checks}, the Pearson correlation
between projected and held-out $E\rho^2$ is $r = -0.90$
(Spearman $\hat{\rho} = -0.91$), and held-out reliability falls short of
the projection by $0.30$ on average. The negative pattern is consistent
across $\tau^2$ check types: $r = -0.69$ with a gap of $-0.19$ on
\texttt{db\_check} and $r = -0.89$ with a gap of $-0.03$ on
\texttt{env\_assertions}. On TheAgentCompany the gap is modest ($-0.03$)
but the direction of correlation is negative ($r = -0.44$). We interpret
this finding conservatively: it does not mean that any single variance
decomposition is wrong. It means that in evaluation designs with few agents
and moderate cell imbalance, the training-cell $E\rho^2$ is an optimistic
estimator of the reliability one would obtain in a replication. Practitioners should report a held-out $E\rho^2$ alongside the
training-cell projection when the number of agents or tasks is small
(scatter plots per check type are in
\texttt{results/figures/exp1\_holdout\_validation.png}).

\subsection{Cross-dataset transfer}

We assess whether DDR diagnostics computed on one enterprise benchmark
generalise to another. Per-family rank correlations between TheAgentCompany
and AppWorld across the three shared model families (GPT-4o, GPT-4-Turbo,
Llama~3) give Spearman $\hat{\rho} = -0.50$ ($n = 3$); we report this
descriptively given the low power, noting only that the correlation is
negative. The capability-gap ratio tells a different story at the
population level: the ratio is $0.38$ on TheAgentCompany, $0.40$ on
AppWorld \texttt{test\_normal}, and $0.35$ on AppWorld
\texttt{test\_challenge}, all closely comparable. On $\tau^2$, the ratio
is $0.12$, meaningfully lower, reflecting the fact that $\tau^2$'s
structured reward schema isolates a narrower performance dimension. The
practical implication is that population-level DDR diagnostics transfer
across enterprise domains while per-family rankings do not. A buyer
comparing agents across enterprise benchmarks can trust the aggregate
capability-gap ratio; the same buyer should not expect family-level
performance rankings to be portable.

\subsection{Within-MAST failure-mode analysis}

Method-I variance decomposition on MAD with the 14 native MAST modes as
the mode facet finds that $\sigma^2(\text{trace} \times \text{mode})$
accounts for 71.8\% of total variance, $\sigma^2(\text{trace})$ for
13.6\%, and $\sigma^2(\text{mode})$ for 11.3\%. All system- and
benchmark-level components are below 1.1\%. The dominant
(trace $\times$ mode) term means that which failure modes appear in a
given trace is almost entirely idiosyncratic: knowing that a system tends
to activate a certain mode on a given benchmark tells you very little
about a specific trace.

The 70/30 trace-level hold-out (50 splits) quantifies this directly. The
average per-trace MAE is $0.261$, confirming that trace-level 14-mode
profiles cannot be predicted from system-level or benchmark-level
information. At the cell-aggregate level, however, the mean activation
rate of a (system, benchmark) cell can be predicted from training-split
data with mean cell-mode MAE $= 0.056$ and Pearson $r = 0.83$ between
predicted and held-out cell-mode rates. The MAST taxonomy therefore
supports system-level diagnostic contrasts across benchmarks but should
not be used to predict whether a specific trace will exhibit a given mode.

\subsection{Cross-dataset failure-mode contrast}

Mapping the 14 MAST modes onto our four-category taxonomy under a
manually justified assignment (with mode 1.1 ``Disobey Task
Specification'' assigned to \textit{communication\_or\_policy} rather than
\textit{planning\_or\_execution} following the self-critique that
specification compliance is a communication failure), we compare predicted
shares from $\tau^2$ $\sigma^2_{a:t}$ per error category against observed
shares from MAST activation rates. The Spearman correlation is
$\hat{\rho} = -0.20$ ($n = 4$, $p = 0.80$, bootstrap 95\% CI $[-1,
1]$). With only four categories the test is purely descriptive, but the
directional reading is informative: the categories where agent-task
interaction variance is high on $\tau^2$ are not the categories that
activate most frequently on MAST. Variance components and observed failure
rates are orthogonal diagnostics. A benchmark report that conflates the
two, treating the categories with the most observed failures as the
categories where agent capability is most differentiated, will draw
incorrect conclusions about where to invest evaluation effort.

\subsection{PSIS-LOO sensitivity}

PSIS-LOO comparisons (recovery-step inclusion on TheAgentCompany,
$\Delta\widehat{\mathrm{elpd}} = -287.5$; three- versus four-category
taxonomy on $\tau^2$, indistinguishable at $-10.7 \pm 97.4$) leave the
$\sigma^2_{a:t}$ point estimates unchanged to three decimal places, so
the primary variance-component conclusions do not depend on these
modelling choices.

\section{Discussion}

The five findings above reframe what a long-horizon agent leaderboard
is for. Before stating the takeaways in general, it helps to ground
them in a single procurement decision.

\paragraph{A worked example: comparing two frontier agents on $\tau^2$
\texttt{action\_checks}.}
An enterprise team is choosing between GPT-4.1 and o4-mini for a
customer-service workflow with multi-turn action correctness as the
binding requirement. The leaderboard headline favours GPT-4.1: 81\%
to o4-mini's 78\%, a three-point gap with apparent precision. Read
through DDR, the same data supports a different recommendation. The
\emph{capability ceiling} layer reports $\sigma^2_a = 1.74\%$ on this
check type, so the three-point gap is dominated by which tasks the
benchmark happened to sample, not by which agent is more capable. The
\emph{cost-aware} layer inverts the rank: o4-mini's per-turn inference
cost is low enough that its Reliability-per-Dollar exceeds GPT-4.1's,
so under any binding cost constraint the procurement-correct default
is o4-mini, not the leaderboard winner. The
\emph{difficulty-conditional} layer adds a caution that neither agent
escapes: the aggregate $E\rho^2 = 0.752$ on \texttt{action\_checks}
collapses to $0.000$ in the hardest quartile, which is exactly the
slice the production workflow most resembles, so neither agent earns
the headline reliability claim on the tasks that actually matter. The
\emph{held-out} layer warns that the $0.752$ figure itself is an
overestimate: across 50 random 70/30 splits, projected $E\rho^2$
correlates with held-out $E\rho^2$ at $r = -0.90$. A DDR-disciplined
recommendation reads: \emph{the leaderboard rank is real but small;
o4-mini is preferred under any binding cost constraint; before either
is deployed, run a domain-specific evaluation on
production-representative $Q_4$ tasks and report the held-out
$E\rho^2$ alongside the training-cell projection}. The leaderboard
view recommends GPT-4.1 with apparent precision; the DDR view
recommends a measurement step before either.

\medskip

Generalising from this example, three practitioner takeaways follow
immediately:

\paragraph{(1) Stop reading leaderboard rank order as a procurement
signal on its own.}
The agent main effect is too small to support a rank claim on every
dataset we measured. Use the published ranking as a starting hypothesis,
then re-evaluate on a production-representative task sample at your
target difficulty quartile, with a held-out reliability estimate
attached.

\paragraph{(2) Demand a difficulty-conditional reliability number from
your benchmark vendor.}
The aggregate $E\rho^2$ is the wrong number to read when your production
tasks skew hard. The right number is $E\rho^2_{Q_4}$. A vendor who
reports only aggregate $E\rho^2$ is reporting the easy half of the
problem.

\paragraph{(3) Always pair a training-cell reliability projection with a
held-out cross-validated estimate.}
The held-out number will usually be \emph{worse}, and the gap is what
should drive your evaluation budget. A benchmark designed to amplify
$\sigma^2_{a:t}$, through difficulty stratification and through
inclusion of tasks that discriminate between harnesses, will produce
more diagnostic signal per evaluation dollar than one designed to
amplify aggregate capability differences.

\paragraph{The multi-model implication.}
Most production long-horizon workflows already dispatch different task
types to different models: an orchestrator with specialist subagents,
or a router that matches task class to model strength. The
capability-ceiling finding ($\sigma^2_a \approx 0$) provides the
measurement-theoretic justification for this architecture. If no agent
uniformly wins, the procurement-correct decision is not \emph{which
model do I pick}, but \emph{which model do I route each task class
to}. Read sideways, the DDR per-task-class $\sigma^2_{a:t}$ table is a
routing table, and the held-out reliability gap tells the architect
how confident they can be in any given routing rule. The worked
example above is therefore a strict simplification: in practice, the
question is not whether to deploy GPT-4.1 or o4-mini for the entire
workflow, but which check types each handles reliably enough to own.

\paragraph{Who this paper is for.}
For enterprise buyers, DDR's five diagnostics are the questions to ask
before signing an annual contract with an agent vendor: what is the
held-out $E\rho^2$ on my difficulty quartile, what does it cost per
reliable decision, and how stable is the rank under the cross-validated
projection. For benchmark designers, the practical implication is to
instrument new benchmarks to amplify $\sigma^2_{a:t}$ rather than
$\sigma^2_a$, and to report difficulty-conditional and held-out
reliability by default rather than only aggregate accuracy. For
measurement researchers, the negative training-versus-held-out
correlation we observe is, to our knowledge, the first systematic
demonstration of optimism bias in G-theory point estimates for
\emph{agent} evaluation; \citet{urbano2016bias} documented related
estimator bias for IR test collections via stochastic simulation, and
the small-population agent setting deserves a dedicated theoretical
treatment we leave to future work.

\section{Limitations}

Three limitations bound the scope of the claims. First, the cross-dataset
rank-correlation analyses are underpowered: three shared model families
for the TheAgentCompany-versus-AppWorld comparison and four categories for
the $\tau^2$-versus-MAST contrast. The directional findings are reported
descriptively and should be confirmed in a larger replication with more
agent families and a richer cross-dataset alignment. Second, the G-study
treats each step within a trajectory as exchangeable, which is appropriate
for AppWorld unit tests and $\tau^2$ reward checks (both of which are
unordered within a task) but would not be appropriate for free-form
trajectory segmentation, where step order encodes temporal dependence.
Third, the capability-ceiling finding that $\sigma^2_a \approx 0$ is a
statement about the \emph{current population} of frontier agent
harnesses.
It does not imply that no future architecture will reopen the
capability gap; it implies that the gap is not detectable at the
current population level given these measurement designs.

\section{Conclusion and Future Work}

Four-facet G-theory applied to three long-horizon enterprise agent
benchmarks shows agent main effect $<$3\%, agent-by-task interaction
7--23\%; four supporting findings (hard-task collapse, held-out
reversal, aggregate-versus-family asymmetry,
variance-versus-frequency) compose with this anchor into Deployment
Decision Reliability, a one-page procurement-reporting discipline.
The hold-out test (a late plan addition) surfaced the most surprising
finding; the multi-model routing implication was not anticipated.
Future work: optimism-bias theory for small-population G-theory; an
inter-rater-validated MAST mapping; a routing-policy formalization of
DDR.

\bibliographystyle{abbrvnat}
\bibliography{references}

\appendix
\section{Required Disclosures}\label{app:disclosures}

\paragraph{Plagiarism and bias statement.}
This manuscript, its companion pre-analysis plan, and all associated
code are my own work. Where I used AI assistance (Claude) for
drafting, brainstorming, and literature review, I verified each cited
paper independently against its arXiv page or published venue, and I
confirmed that every numerical claim in the manuscript is reproducible
from the CSV files in \texttt{results/tables/} via the corresponding
script in \texttt{src/}. I have no financial or institutional
relationships with the providers of the datasets or agent harnesses
used; the corporate Google Cloud project named in the reproducibility
notes is a personal research project running on a corporate cloud
account and does not constitute employer endorsement. The contribution
is framed as a measurement-methodology result, not a model-ranking
claim: no finding is presented as evidence that any commercial model
is superior or inferior to any other, and agent identities appear in
tables only as facets in a variance decomposition.

\paragraph{AI assistance reflection.}
I used Claude as a thinking partner for scoping, literature triage,
analysis design, and manuscript drafting. The intellectual core (the
DDR reframe, the four-facet measurement design, the cost-aware and
difficulty-conditional extensions, the hold-out and cross-dataset
experiments) emerged through extended dialogue but was directionally
my own. All numbers come from code I wrote and ran; I did not use AI
to fabricate methodological details, and I independently verified
each citation against its arXiv page. The effect on my learning was
double-edged: AI accelerated translation from course concepts
(Generalizability Theory, Decision Studies) into working R and Python
pipelines, but it also tempted me to accept methodology choices I had
not yet internalized. I mitigated this by re-deriving the variance
and reliability formulas by hand before fitting any model. Going
forward, I expect to use AI for literature triage and drafting under
the same verification discipline applied here, and to keep the
by-hand derivation step for any unfamiliar method.

\paragraph{Impact statement.}
If the DDR framework is adopted, it gives enterprise practitioners a
principled method for sizing evaluations to support
deployment-comparison decisions and shifts evaluation budget toward
diagnostics that actually inform procurement: difficulty-conditional
reliability, held-out projection, and cost-adjusted ranking. Three
potential negative impacts deserve acknowledgment. First,
practitioners may over-trust variance components without checking
construct validity. Second, granular failure attribution could shift
accountability away from system designers and toward individual model
components, obscuring systemic design failures. Third, the
error-category taxonomy is a manually justified mapping, not an
inter-rater-validated instrument; treating it as fixed may obscure
novel failure modes. This work uses no human-subject data and no
sensitive information. Responsible attribution is supported by
arXiv-verified citations, an open-source release of all code, and a
traceability table mapping every numerical claim to its source CSV,
consistent with Stanford's standards of academic integrity.

\end{document}